# Structured Pruning of Deep Convolutional Neural Networks

**Sajid Anwar, Kyuyeon Hwang and Wonyong Sung**
Department of Electrical Engineering and Computer Science
Seoul National University
Seoul 151-744, Korea
*{Sajid, khwang}@dsp.snu.ac.kr, wysung@snu.ac.kr*

## Abstract

Real time application of deep learning algorithms is often hindered by high computational complexity and frequent memory accesses. Network pruning is a promising technique to solve this problem. However, pruning usually results in irregular network connections that not only demand extra representation efforts but also do not fit well on parallel computation. We introduce structured sparsity at various scales for convolutional neural networks, which are channel wise, kernel wise and intra kernel strided sparsity. This structured sparsity is very advantageous for direct computational resource savings on embedded computers, parallel computing environments and hardware based systems. To decide the importance of network connections and paths, the proposed method uses a particle filtering approach. The importance weight of each particle is assigned by computing the misclassification rate with corresponding connectivity pattern. The pruned network is re-trained to compensate for the losses due to pruning. While implementing convolutions as matrix products, we particularly show that intra kernel strided sparsity with a simple constraint can significantly reduce the size of kernel and feature map tensors. The pruned network is finally quantized with reduced word length precision. This results in significant reduction in the total storage size providing advantages for on-chip memory based implementations of deep neural networks.

## 1    Introduction

Convolutional neural networks (CNN) have been successfully applied to several diverse classification problems including speech and image recognition [1][2][3]. In CNN design, we need to decide the optimum network architecture and parameters count for a specific task. Large networks have the capacity to learn more difficult functions at the cost of increased computational complexity. However if the current parameters count is greater than the unknown optimum number, overfitting may occur. On the other hand, too few parameters limit the network's learning capability. One efficient approach for training is to learn a task with a large sized network and prune it by removing redundant and duplicate connections. This results in comparable level of performance with fewer parameters and better generalization [4]. Another important problem is porting deep learning algorithms to resource limited portable devices. With the advances in deep learning, it can make the smartphones and other machines even smarter. For this purpose, researchers have proposed ideas for designing the best performing lightweight networks. Lightweight neural networks have improved generalization and the parameters can be saved in on chip memory. This results in energy savings as frequent DRAM accesses consume much energy. Since deep neural networks conduct many multiply and accumulate (MAC) operations, sparsity helps in reducing these computations. However, the computational complexity of a neural network depends on not only the number of parameters or arithmetic operations but the architecture, layer types and connectivity patterns. Irregular sparsity is difficult to exploit for efficient computation. The proposed work investigates and proposes ideas to solve such problems with structured pruning and fixed point optimization.

Pruning is useful in several ways. First, gradual network pruning inherits knowledge from the bigger and cumbersome network. Secondly, directly learning a complex function with lightweight network may not yield acceptable results. Moreover, in practice, the optimum network architecture is unknown. In the literature large sized deep networks [1][2][3] have achieved state-of-the-art performance on various challenging tasks. It is therefore appropriate to first learn a task with many parameters followed by pruning redundant and less important connections.

Pruning techniques can be broadly categorized as structured or unstructured. Unstructured pruning does not follow a specific geometry or constraint. In most cases, this technique needs extra information to represent sparse locations. It depends on sparse representation for computational benefits. On the other hand, structured sparsity places non-zero parameters at well-defined locations. This kind of constraint enables modern CPUs and graphics processing units (GPUs) to easily exploit computational savings. In this work, we explore channel, kernel, and intra-kernel sparsity as a means of structured pruning. In channel level pruning, all the incoming and outgoing weights to/from a feature map are pruned. Channel level pruning can directly produce a lightweight network. Kernel level pruning drops a full $k \times k$ kernel, whereas the intra-kernel sparsity prunes weights in a kernel. The proposed work shows that when combined with convolution lowering [5] [6], the intra kernel strided sparsity can significantly speedup convolution layer processing. The kernel level pruning is a special case of intra-kernel sparsity with 100% pruning. These granularities can be applied in various combinations and different orders.

Network pruning has been studied by several researches [7]-[12]. The works of [7][8] have shown that a much bigger portion of weights can be set to zero with minimum or no loss in performance. They train a network with L1/L2 norm augmented loss functions and gradually prune it. If the value of a connection is less than a threshold, the connection is dropped. The authors in [7] further extend this work by quantizing the finally pruned network [8]. However [7] and [8] have to use sparse representation to benefit from the induced sparsity. Units in the hidden layers are pruned in [9] for a feed-forward deep neural network. Computational complexity is reduced with sparse connectivity in convolution and fully connected layers in [10]. [11] prunes multi-layered feed forward networks with genetic algorithm and simulated annealing. A survey on pruning techniques is reported in [12]. These works [9]-[12] utilize unstructured sparsity in feed forward neural networks. A recently published work induces channel wise sparsity in a network [13]. Compared to [13], the proposed work explores sparsity at multiple levels using an efficient search followed by fixed point optimization. Dropout [14] and Dropconnect [15] zeroes neuron outputs and weights only during training and the network architecture does not change at evaluation time. Both techniques train different subsets of network parameters during training which results in better generalization. Our work drops parameters permanently and yields network with fewer parameters at test time. In [5][6], convolutions are converted to matrix-matrix multiplication, which follows from the same logic that two bigger sized matrix multiplications are better than several small sized ones [6]. Fixed-point optimization is used to reduce the memory and computational requirements of deep networks [16] [17].

Another contribution of the proposed work is the use of particle filter to locate pruning candidates. We search for the most likely connection combinations and prune the rest. Each particle simulates one such set of connections or masks. We also apply fixed point optimization to reduce the word length of network parameters. We represent weights and signals with 4 or 5 bits precision while maintaining the same level of performance. This further reduces the storage requirement and is advantageous for on-chip based implementation of deep learning algorithms.

The rest of the paper is organized as follows. Section 2 briefly introduces CNN in the context of pruning. The pruning criterion is outlined in Section 3. This section also discusses pruning granularities, particle filter, genetic algorithm and the hybrid evolutionary particle filter. Section 4 explains the iterative pruning process and fixed point optimization. Experimental results and discussions are provided in Section 5. Finally Section 6 introduces the future work and concludes.

## 2      Convolutional Neural Network

This section briefly introduces CNN in the context of pruning. CNN was first proposed and applied to handwritten digit recognition by Lecun et.al. [18]. CNN has more diverse layer types than deep neural network (DNN). It has convolution, pooling and fully connected layers. The convolution and pooling layers act as feature extractors. At the rear end, near the output layer, CNN employs fully connected layers for classification. A sample CNN network is shown in Fig. 1. In CNN, the most computationally expensive layers are the convolution layers. The fully connected layers are implemented as matrix multiplications. Convolution layer convolves the $k \times k$ kernels with $n$ feature maps of previous layer. If the next layer has $m$ feature maps, then $n \times m$ convolutions are performed and $n \times m \times (H \times W \times k \times k)$ MAC operations are performed, where $W$ and $H$ represents the feature map width and height of the next layer. Further the convolution layers default memory access pattern is not cache friendly. It is therefore highly desirable to reduce the complexity of convolution layers. In the literature, there have been various attempts to reduce the computational complexity of convolution layers. The work of [5] and [6] converts convolutions into matrix-matrix multiplication. This avoids the usage of 4 to 6 levels of nested loops and speeds up the computation by 3 times [5]. However redundant data and kernels storage has its own cost of extra memory usage. Our work proposes to reduce this complexity with structured pruning and fixed point optimization. We show that strided sparsity is helpful in reducing the size of matrices in convolution lowering [5]. The next section discusses this in detail.

## 3      Neural Networks Pruning

Pruning permanently drops less important connections from a network for computational benefits. Unstructured pruning requires sparse representation schemes for reducing computations, which however demands addressing overhead for computing addresses of non-zero elements. Structured pruning has no or little extra cost and can be easily exploited for efficient implementation. We introduce structured pruning at various granularities. In this section, we present these granularities and the selection of pruning candidates with an evolutionary particle filter (EPF).

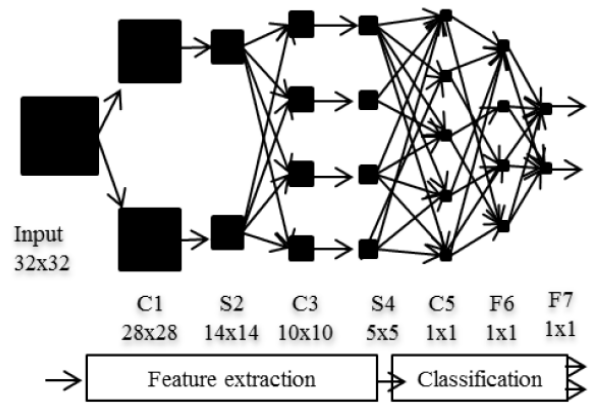

**Figure 1 Convolutional neural network with eight layers. Layers C1, C2 and C3 are the convolution layers while S2 and S4 constitute the two pooling layers. F6 and F7 are the two fully connected layers. This network can be represented with a string of 1-2-2-4-4-5-4-2 where each number denotes the count of feature maps in that layer [16].**

## 3.1 Pruning Granularities

Taking computational advantages using randomly scattered unstructured sparsity in a network is very difficult. It demands many conditional operations and extra representation to denote the location of zero or non-zero parameters. Generally, the convolution layers in the non-pruned network have fully connected convolution connections. In Fig. 2, the layer $L_1$, $L_2$ and $L_3$ contain 2, 3, and 3 feature maps, respectively. As channel and feature map are similar concepts, we therefore used them interchangeably throughout this article. The number of convolution connections between $L_1$ and $L_2$ is 6 (=2×3) and that between the $L_2$ and $L_3$ are 9. Each feature map in $L_2$ has a $K \times K$ convolution connection from each channel in $L_1$. Thus, the pruning exploiting the largest granularity is deleting a feature map or feature maps. If a feature map in a layer is removed, all the incoming and outgoing kernels are pruned. Fig. 2 shows 5 pruned kernels with a red dashed line. Considering the configuration in Fig. 2, the 2-3-3 architecture is reduced to 2-2-3.

The next level pruning is deleting kernels where each kernel represents one whole convolution. The kernel level sparsity is depicted with blue dotted lines in Fig.2.

The lowest level pruning is using the intra-kernel sparsity, which forces some weights into zero valued ones. In the previous works, the intra-kernel level pruning is usually conducted by zeroing small valued weights [7] [8]. We particularly explore the intra-kernel level pruning using the sparsity at well-defined locations, which is called the intra kernel strided sparsity. Fig. 2(b) depicts this idea. The starting index for the first non-zero element is randomly assigned. Therefore the strided sparsity associates an offset as the starting index and the stride size with each kernel. Figure 3 shows how the strided sparsity and a simple constraint can reduce the size of feature and kernel matrices. The intra kernel strided pruning has the potential to bridge the gap between pruning and its computational advantages. Combined with convolution lowering, it can significantly reduce the computational cost.

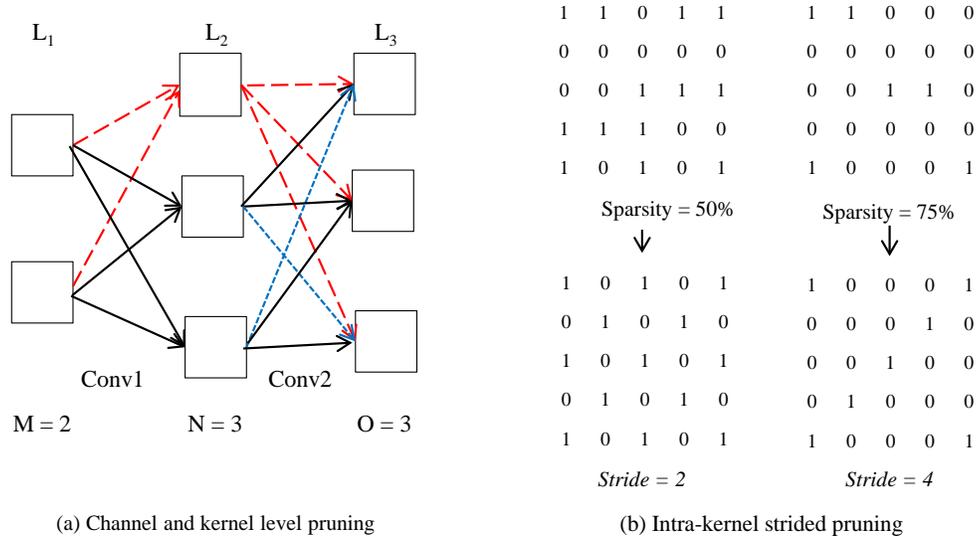

(a) Channel and kernel level pruning     (b) Intra-kernel strided pruning

**Figure 2 (a) shows channel and filter wise pruning. The red dashed line shows channel level pruning. When we prune all the incoming filters to a feature map, all the outgoing kernels are also pruned. The blue dotted line depicts pruning *kxk* kernels. (b) Shows intra kernel level sparsity for both structured and unstructured cases. Kernel level pruning (blue dotted) is a special case of intra-kernel pruning, when the sparsity rate is 100%.**

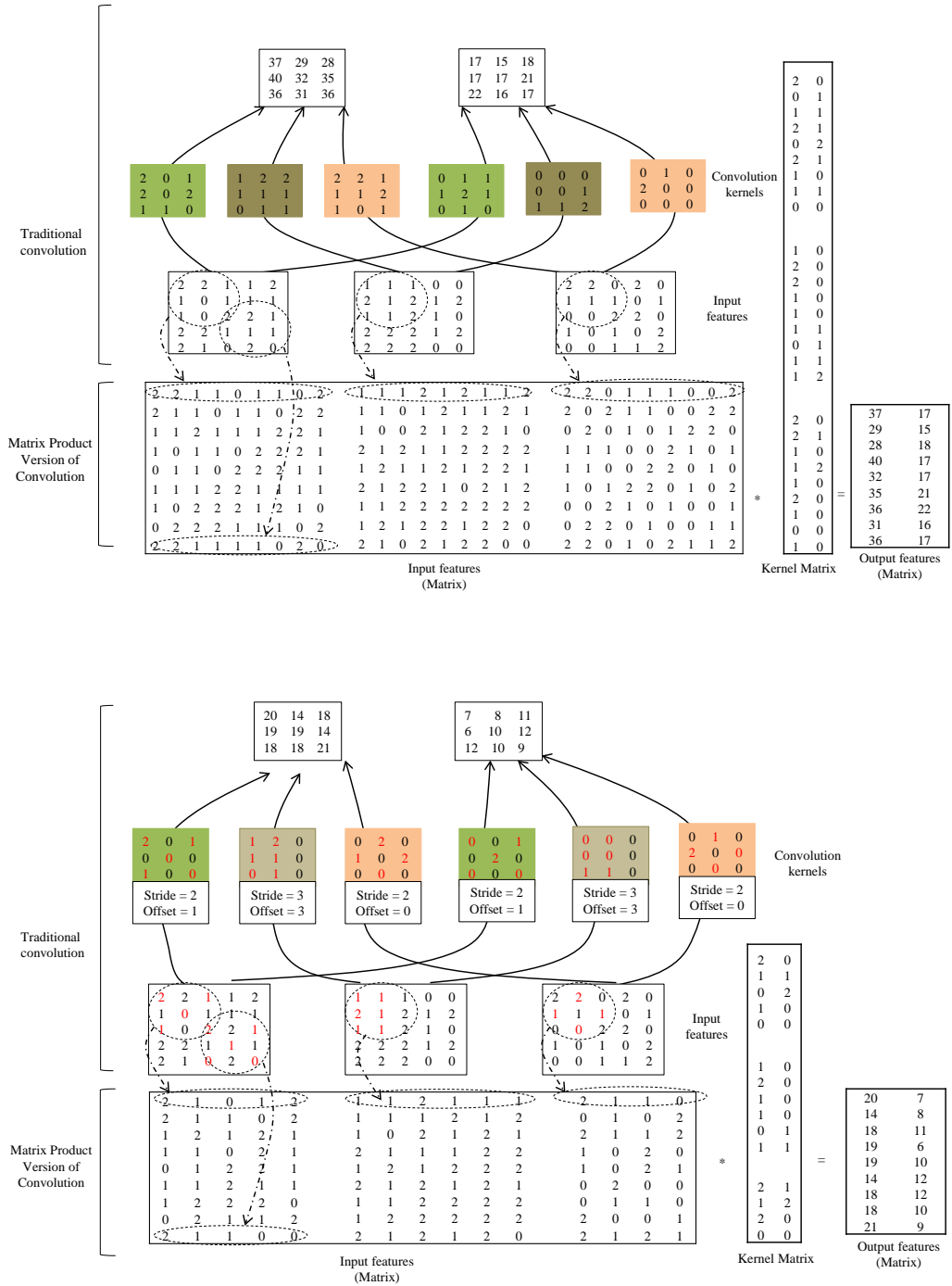

**Figure 3** The top figure provides an example of convolution lowering idea introduced in [5][6]. (b) Our proposed idea constrains each outgoing convolution connection for a source feature map to have similar stride and offset. The offset shows the index of first pruned weight. The constraint is shown with the similar colored background squares. This significantly reduces the size of both features matrix and kernel matrix. The first 9 columns in row 1 of the input feature matrix changes from 2 <u>2</u> 1 <u>1</u> 0 <u>1</u> 1 <u>0</u> 2 to 2 1 0 1 2 with the underlined elements pruned. Only the red colored elements in the feature maps and kernels survive and the rest are pruned. For this example, the size of feature matrix is reduced from 9×27 to 9×15 and the kernel matrix size is reduced from 27×2 to 15×2. (Better seen in color)

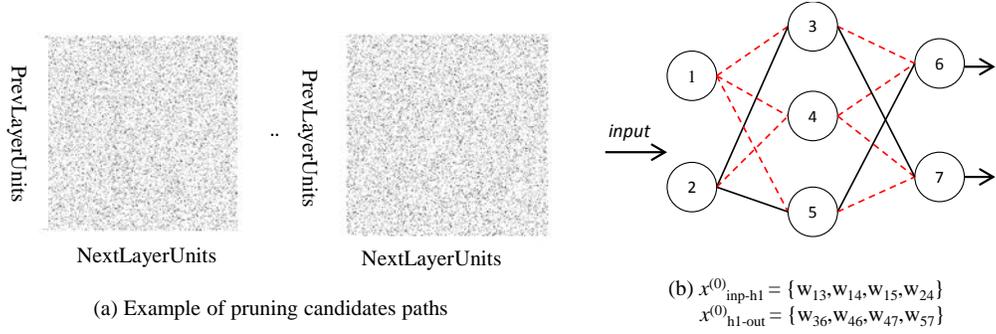

(a) Example of pruning candidates paths

(b) $x^{(0)}_{\text{inp-h1}} = \{w_{13}, w_{14}, w_{15}, w_{24}\}$
$x^{(0)}_{\text{h1-out}} = \{w_{36}, w_{46}, w_{47}, w_{57}\}$

**Figure 4 (a)** shows the dotted matrices as the pruning masks for weights between two layers. The *x* state vector represents this mask. In (b), one example of the state vector is provided. The circles represent the neurons while $w_{ij}$ shows the weight going from neuron i to j.

### 3.2 Particle Filter and Evolutionary Particle Filter

The pruning process needs to select less important connection combinations as pruning candidates. These connections, when pruned, have least adversary on the network performance which can be compensated with re-training. Further the highly likely connections survive through iterative pruning. In this work, we propose to locate pruning candidate connections with the sequential Monte Carlo (SMC) method also known as particle filters [19]. Particle (PF) filters finds its applications in several fields [19]-[22]. With a set of weighted particles, the PF represents the filtering distribution [23]. Particle filter is usually applied to the system model shown in equation (1) and (2). The state vector is represented by *x*, *k* shows the time step, $z_k$ is the observation vector, $v_k$ is observation noise and $\mu_k$ is the process noise. The observation function is represented by *h(.)* whereas *f(.)* represents the transition function. When pruning *n* connections, the possible combinations are on the order of $O(2^n)$ which means that exhaustive search is not feasible. With *N* particles, we simulate several possible connection combinations. Fig. 4. shows an example of particle's state vector. The trained network is used as the observation function which is noisy as the classification error rate is greater than 0%. We evaluate the misclassification rate (MCR) on the evaluation set for each particle. This way importance weight is computed by 1 – MCR. Connections with high importance survive through several iterations while the rest are pruned. MCR compares the network assigned label with the true label which guides the network learn the true labels. Once all particles are assigned there probabilities, then we construct the cumulative distribution function (CDF) and resample with the sequential importance resampling (SIR) [19]. The transition function is simulated by perturbing the pruning mask.

$$\mathbf{x_k} = f(\mathbf{x_{k-1}}) + \mathbf{\mu_k} \qquad (1)$$

$$\mathbf{z_k} = h(\mathbf{x_k}) + \mathbf{v_k} \qquad (2)$$

PF with finite number of samples suffer from degeneracy and impoverishment problems when less likely particles are replaced by highly likely particles [24]. The evolutionary particle filter (EPF) proposes a hybrid approach where genetic algorithm (GA) is combined with PF [25]. Particles are similar to chromosomes and survival of the fittest has equivalence to the resampling algorithm. With the hybrid approach, the aim is to increase the fitness of the whole population. The GA augmented approach re-supplies or re-defines particles in less likely regions while maintaining the highly likely genes in its chromosome. EPF reduces computational cost as it requires fewer particles than conventional particle filters [25]. To help reducing the cost of finding pruning candidates, the importance weight assignment uses the small sized evaluation set. We also consider that SMC techniques have more potential usages in exploring the network parameters space.

## 4     Network Retraining and Fixed Point Optimization

The proposed work first trains a network to the baseline. This is followed by pruning which reduces the effective number of network parameters. Pruning degrades the network performance which is compensated by re-training the pruned network. We set the pruning limit for each layer depending on its parameter count and learning capacity. Usually, the first convolution layer has fewer parameters than the following layers. Secondly it directly performs on the input layer and is therefore more sensitive to pruning [7]. We therefore set a lower limit for the first convolution layer. During the whole process, we set aside validation set for network convergence. We particularly show a plot where sparsities at various levels are applied with different combinations.

In this work we further reduce the memory requirement by quantizing the pruned network with fixed-point optimization. Quantization techniques are orthogonal to network pruning techniques and hence can supplement each other [8]. The fixed-point optimization algorithm is outlined in [16] [17]. During fixed-point optimization, the network keeps both high (32-bit) and low precision weights. The algorithm obtains quantized weights from floating points using L2 error minimization. When grouping weights for quantization, each convolution kernel has one quantization step size. The quantized weights are used in the feed forward path whereas the corresponding floating point weights are updated in the backward path. The algorithm in [17] [27] also evaluates layer wise sensitivity analysis and signal quantization. We apply these techniques to the pruned network to further squeeze the memory and computational requirements. The next section provides the initial experimental results and discussions.

## 5     Experimental Results

We first show that pruning enables a network to learn better with fewer parameters and resources. Fig. 6 shows the results of our first experiment. We experiment with two CNN architectures: NW1 = 3-32-32-32-32-64-10 and NW2 = 3-32-32-21-21-38-10. NW2 has about half convolution connection of NW1 (1566/3168). We train NW1 on CIFAR10 dataset and achieve the baseline result shown as black solid line in Fig. 6. This trained network is then pruned to obtain NW2 which is shown as green line in Fig. 6. We can find that the performance is quite close to the baseline. In the second case, we randomly initialize NW2 and train it on CIFAR-10. This is shown as red dashed line in Fig. 6.  We can observe that this network cannot reach close to the baseline or the performance of the pruned network. This indicates that pruning helps the network learn better. Further transfer learning enables the lightweight network to inherit useful knowledge from the bigger sized predecessor. For this experiment, we only conduct channel and kernel level pruning in the second (32 – 32) and the third (32 – 64) convolution layers. Fig. 8 shows a mask matrix at an intermediate stage of pruning. Vertical black lines show the pruning of feature maps in the destination layer while the horizontal black lines show pruning of source feature maps. Also kernel level pruning appears as black squares in Fig. 8 (c). It can be observed that the sparsity is well structured. We present experimental results on two datasets: MNIST [18] and CIFAR-10 [28]. We observe that pruned networks show better performance than the non-pruned one because of generalization. However when we drop too many parameters, the network capability is reduced and that's where the error rate starts rising up.

The strided pruning is useful as it introduces intra kernel structured sparsity. We are conducting experiments on evaluating its effectiveness for convolution lowering [5].The proposed work is unique in the sense that it is friendlier for computational benefits. During training and pruning, we use stochastic gradient descent (SGD) with a mini-batch size of 128 and RMSProp [26].

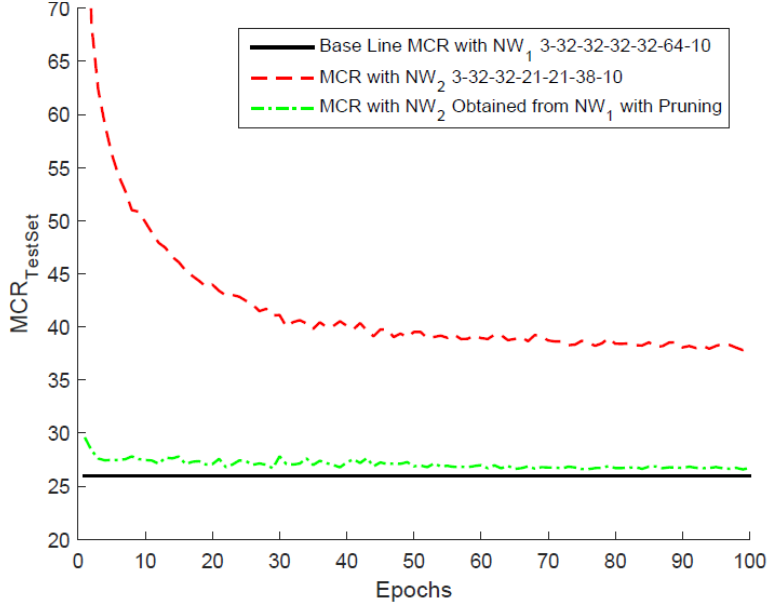

**Figure 6 This plot shows that directly training a small sized network cannot reach the performance level of a similar sized network obtained through pruning.**

### 5.1 CIFAR-10

The CIFAR-10 dataset consists of a ten class classification problem [28]. The dataset includes samples from ten classes: airplane, automobile, bird, cat, deer, dog, frog, horse, ship and truck. The training set consists of 50,000 RGB samples. Test set contains 10,000 samples. Each sample has 32x32 resolution. We first conduct the experiment with a CNN network having 3-32-32-32-32-64-10 architecture. We divide the training set into the validation set (5000) and training set (45000). We train the network with rectified linear units and SGD. The network has a total of 79978 parameters. The channel wise sparsity reduces the network size to 3-32-32-21-21-38-10. The first convolution layer is not pruned. For the $2^{nd}$ convolution layer, the numbers of convolution connections are reduced to 672/1024 while the $3^{rd}$ one is reduced to 798/2048. As a result 35% and 62% of the convolution connections are dropped with less than 1 % increase in MCR. The first stage had a total of 50% sparsity. Increasing channel level sparsity beyond this point causes an increase in the MCR by more than 1%.

In the next set of experiments, we evaluated sparsities in various combinations. Fig.7. summarizes the results. We can find that channel level pruning is very destructive as it affects a large number of parameters. Intra kernel sparsities can be induced in various proportions. For example in a *kxk* kernel, weights can be pruned with a ratio of 1/2 or 2/3 etc. Our experimental results show that we can reduce the size of the network by 75% with minimum loss in accuracy. This is achieved with macro pruning (46%) followed by intra-kernel strided sparsity (shown in Fig. 7 with a black solid line extending from the red solid line). We further emphasize that this kind of sparsity is well structured and can be directly encoded in representing convolutions as matrix-matrix multiplication.

We also apply fixed point optimization to the finally pruned network. The results are shown in Table 1. We observe that the pruned network can be represented with 3 or 4 bit precision. Here, the precision of 31-levels denotes 5 bit representation. This further reduces the network size by a significant amount for on-chip memory based VLSI implementations.

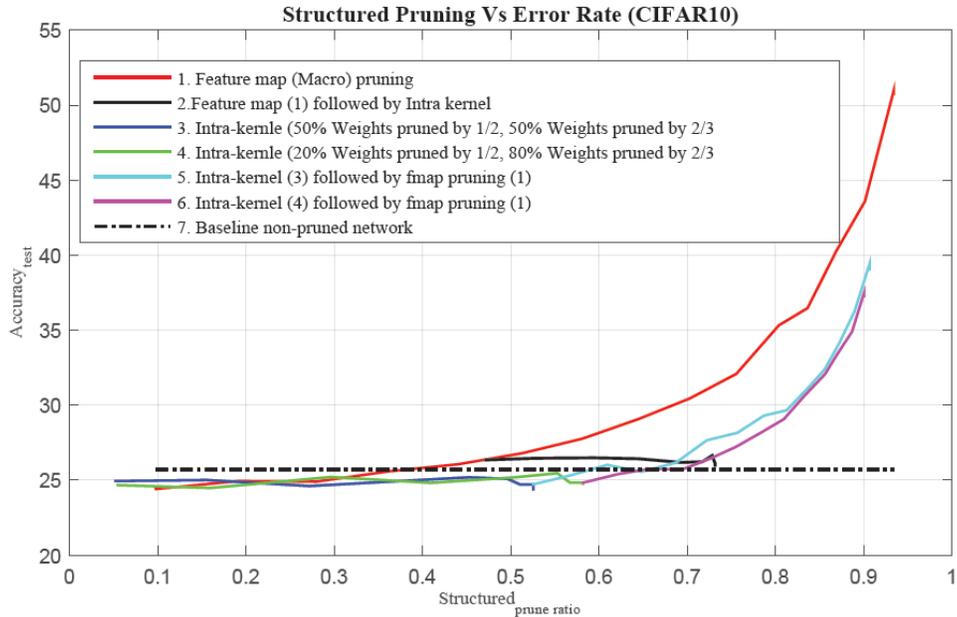

**Figure 7.** The plot shows pruning applied in various combinations. It is observed that feature map/channel level pruning followed by intra kernel pruning provides the best result.

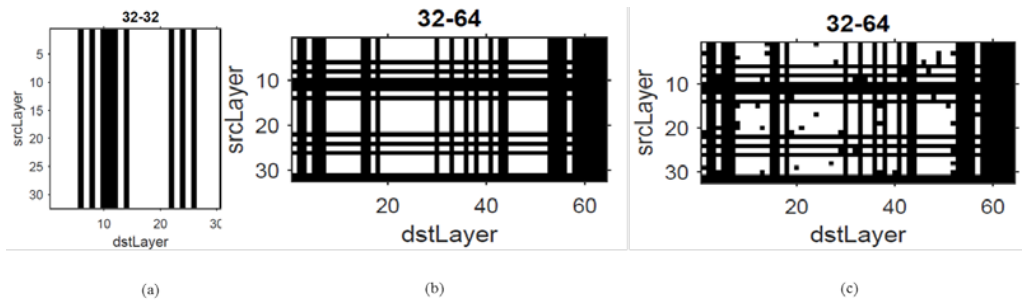

**Figure 8** shows channel and kernel level sparsity. Each entry in the matrix represents a kernel (a) shows channel level sparsity of 32-32 layer. In this layer we have a total of 1K convolutions. The figure shows that this number is reduced to 32×24. This also causes kernel level pruning in the next convolution layer, which is shown in (b) as horizontal black lines. (c) The black squares show kernel level pruning.

**Table 1 CIFAR-10 fixed point optimization of pruned NW**

| Convolution Layers (M Levels) | | | Rear Layers (M Levels) | MCR |
|---|---|---|---|---|
| C1 3x32 | C3 32x32 | C5 32x64 | F6 | Test Set MCR (%) |
| 3 | 3 | 3 | 7 | 31.81 |
| 7 | 7 | 7 | 15 | 27.80 |
| 15 | 15 | 15 | 15 | 27.06 |
| 31 | 31 | 31 | 31 | 26.21 |

## 5.2 MNIST

MNIST is a handwritten digit recognition dataset consisting of 70,000 grey scale images of 28x28 resolutions [18]. These images are divided into training (60,000) and test (10,000) sets. We first evaluate the effectiveness of structured sparsity for the network architecture 1-20-20-50-50-500-10 [7] [18]. This architecture has two convolution and two pooling layers. The first convolution layer has *1x20x5x5* weights, whereas the second convolution layer has *20x50x5x5* weights. We first train the network using 50,000 training samples and 10,000 validation samples with SGD. We obtain 0.80% test error rate after training. This is followed by pruning the network. We do not prune the first layer as it has only twenty convolution connections. Feature map/channel level pruning is applied to the second convolution layer. At 10th pruning iteration, the network size is reduced to 1-20-20-20-20-500-10 with 0.93% MCR on the test set. The number of convolution connections in the 2nd layer has been reduced to 400/1000. This is a reduction by 60% in the second convolution layer. This has an indirect impact on the first fully connected layer which reduces to 20(4x4)-500 from 50(4x4)-500. It is important to recall that this kind of sparsity does not demand extra representational efforts and the sparsity is well structured and directly reduces computational cost.

## 5 Conclusion

In this work we explored structured sparsity in deep convolutional neural networks. The sparsity in channel and kernel level is explained along with the intra-kernel strided one. We found that channel level sparsity cannot be applied in higher proportion as beyond some limit it affects the network's representational capacity. Further we found that intra-kernel strided sparsity along with convolution lowering can significantly reduce the computational complexity of convolutions. Quantization and pruning are orthogonal techniques and can augment the savings. In our future work, we are exploring other related aspects of this work and profiling the execution time effects due to reduced size convolution lowering.